\documentclass{article}
\pdfoutput=1

\usepackage{PRIMEarxiv}

\usepackage[utf8]{inputenc} 
\usepackage[T1]{fontenc}    
\usepackage{hyperref}       
\usepackage{url}           
\usepackage{booktabs}       
\usepackage{amsfonts}      
\usepackage{nicefrac}       
\usepackage{microtype}      
\usepackage{lipsum}
\usepackage{fancyhdr}       
\usepackage{graphicx}       
\graphicspath{{media/}}     
\usepackage{multirow} 
\usepackage{bbding} 
\usepackage{caption} 
\usepackage{subcaption} 

\title{Facial Expression Recognition with Swin Transformer}

\author{
  JUN-HWA KIM \\
  Department of Electronics and Electrical Engineering \\
  Dongguk University \\
  Seoul, Korea\\
  \texttt{jhkim414@dongguk.edu} \\
    \And
  NAMHO KIM \\
  Department of Electronics and Electrical Engineering \\
  Dongguk University \\
  Seoul, Korea\\
  \texttt{namho96@dgu.ac.kr} \\
   \And
  CHEE SUN WON \\
  Department of Electronics and Electrical Engineering \\
  Dongguk University \\
  Seoul, Korea\\
  \texttt{cswon@dongguk.edu} \\
}

\begin{document}
\maketitle

\begin{abstract}
The task of recognizing human facial expressions plays a vital role in various human-related systems, including health care and medical fields. With the recent success of deep learning and the accessibility of a large amount of annotated data, facial expression recognition research has been mature enough to be utilized in real-world scenarios with audio-visual datasets. In this paper, we introduce Swin transformer-based facial expression approach for an in-the-wild audio-visual dataset of the Aff-Wild2 Expression dataset. Specifically, we employ a three-stream network (i.e., Visual stream, Temporal stream, and Audio stream) for the audio-visual videos to fuse the multi-modal information into facial expression recognition. Experimental results on the Aff-Wild2 dataset show the effectiveness of our proposed multi-modal approaches.
\end{abstract}


\section{Introduction}
In recent years, the recognition of human facial expressions becomes a very popular task not only in Artificial Intelligence (AI) research but also in practical applications such as health care and medical fields \cite{muhammad2017facial,davoudi2019intelligent}. Especially, the remarkable advances in deep learning and the availability of massive annotated datasets pave the way for real-world scenarios of facial expression recognition. In response to this trend, the 3rd Affective Behavior Analysis in-the-wild (ABAW 2022) competition by Kollias et al. \cite{kour2014real, kour2014fast,hadash2018estimate,kollias2022abaw, kollias2021analysing, kollias2020analysing, kollias2021distribution, kollias2021affect, kollias2019expression, kollias2019face, kollias2019deep, zafeiriou2017aff} in conjunction with CVPR 2022 provides a large-scale in-the-wild dataset of Aff-Wild2. The Aff-Wild2 consists of 548 videos with 2,813,201 frames and provides annotations for three main tasks of valence-arousal estimation, action unit (AU) detection, and eight facial expression classification. Valence indicates how positive the person is, and arousal indicates how active this person is. AUs are the basic motions of an individual or muscle group to describe an emotion. The eight facial expressions include neutral, anger, disgust, fear, happiness, sadness, surprise, and other.

In this paper, we deal with the eight facial expression classifications. To tackle the problem, we propose a three-stream network that utilizes multi-modal information, such as spatial, temporal, and audio information. The three-stream network is divided into the visual stream, the temporal stream, and the audio stream. The visual stream uses a single frame, while the temporal stream makes use of multiple frames. The audio stream uses an image obtained by converting an audio signal into a mel-spectrogram.

\section{Methodology}

\subsection{Three stream network}

\begin{figure}[h]
    \includegraphics[width=15cm]{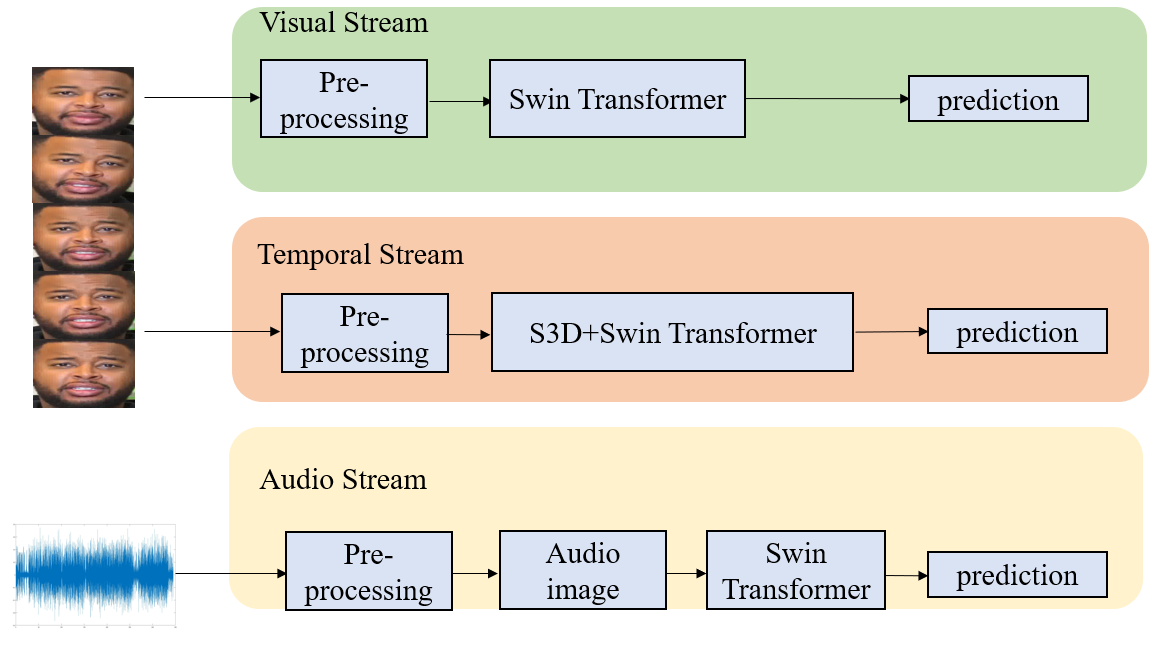}
    \caption{Overall structure.}
    \label{network}
\end{figure}

The proposed three-stream network is to perform the multi-modal task of Aff-Wild2. Since Aff-Wild2 provides cropped face images and audio extracted from video frames, facial expression needs to be predicted comprehensively based on these multi-modal data. The proposed three-stream network is shown in Figure \ref{network}, which is composed of a visual stream, a temporal stream, and an audio stream. Swin transformer \cite{liu2021swin} is used as the backbone of all streams, and the S3D (Shallow 3D CNN) \cite{kim2021deep} structure is used in the visual stream-video shot that processes multiple frames. We independently trained each visual stream-image, visual stream-video shot, and audio stream, and then performed score fusion in the final inference. In the pre-processing stage of the visual stream, resize, augmentation and frame sampling of the input are performed. In the pre-processing stage of the audio stream, the audio signal is converted into a mel-spectrogram, and resize and augmentation are performed.

\subsection{Dataset}

The Aff-Wild2 \cite{kollias2019expression} data set consists of 564 videos and 2.8M frames. The Aff-Wild2 basically consists of 8 classes with 7 emotions (’Neutral’, ’Anger’, ’Disgust’, ’Fear’, ’Happiness’, ’Sadness’, ’Surprise’) and ’Other’ classes. Among them, the training data consists of 253 videos and the validation data consists of 70 videos. Table \ref{table_dataset_stat} shows the number of images and imbalance ratio to total for 8 emotions of Aff-Wild2.

\begin{table*}[h]
  \caption{Image distribution among expression classes for the Aff-Wild2 dataset.}
  \centering
  \begin{tabular}{ccccccccccc}
    \toprule
    \multicolumn{2}{c}{\multirow{2}{*}{Dataset}} & \multicolumn{8}{c}{Expression} & \multirow{2}{*}{Total} \\ \cmidrule{3-10}
    {} & {} & Neutral & Anger & Disgust & Fear & Happiness & Sadness & Surprise & Other & {} \\
    
    \midrule
    \multirow{4}{*}{Train} & Number & \multirow{2}{*}{175500} & \multirow{2}{*}{16356} & \multirow{2}{*}{10725} & \multirow{2}{*}{9080} & \multirow{2}{*}{89917} & \multirow{2}{*}{79140} & \multirow{2}{*}{30096} & \multirow{2}{*}{163189} & \multirow{2}{*}{574003}\\ 
    &  of Data & \\
    \cmidrule{2-11}
    &  Ratio to & \multirow{2}{*}{0.306} & \multirow{2}{*}{0.028} & \multirow{2}{*}{0.019} & \multirow{2}{*}{0.016} & \multirow{2}{*}{0.157} & \multirow{2}{*}{0.138} & \multirow{2}{*}{0.052} & \multirow{2}{*}{0.284} & \multirow{2}{*}{1.000}\\
    &  total & \\
    \midrule
    \multirow{4}{*}{Validation} & Number & \multirow{2}{*}{82099} & \multirow{2}{*}{6056} & \multirow{2}{*}{10725} & \multirow{2}{*}{8388} & \multirow{2}{*}{33581} & \multirow{2}{*}{24633} & \multirow{2}{*}{12302} & \multirow{2}{*}{103833} & \multirow{2}{*}{281617} \\
    &  of Data & \\
    \cmidrule{2-11}
    &  Ratio to & \multirow{2}{*}{0.292} & \multirow{2}{*}{0.022} & \multirow{2}{*}{0.038} & \multirow{2}{*}{0.030} & \multirow{2}{*}{0.119} & \multirow{2}{*}{0.087} & \multirow{2}{*}{0.044} & \multirow{2}{*}{0.369} & \multirow{2}{*}{1.000} \\
    &  total & \\
    \bottomrule
  \end{tabular}
  
  \label{table_dataset_stat}
\end{table*}

\subsection{Half-mix jittering: New Data Augmentation for Single Image}

\begin{figure}[h]
     \centering
     \begin{subfigure}[b]{0.3\textwidth}
         \centering
         \includegraphics[width=\textwidth,height=7cm]{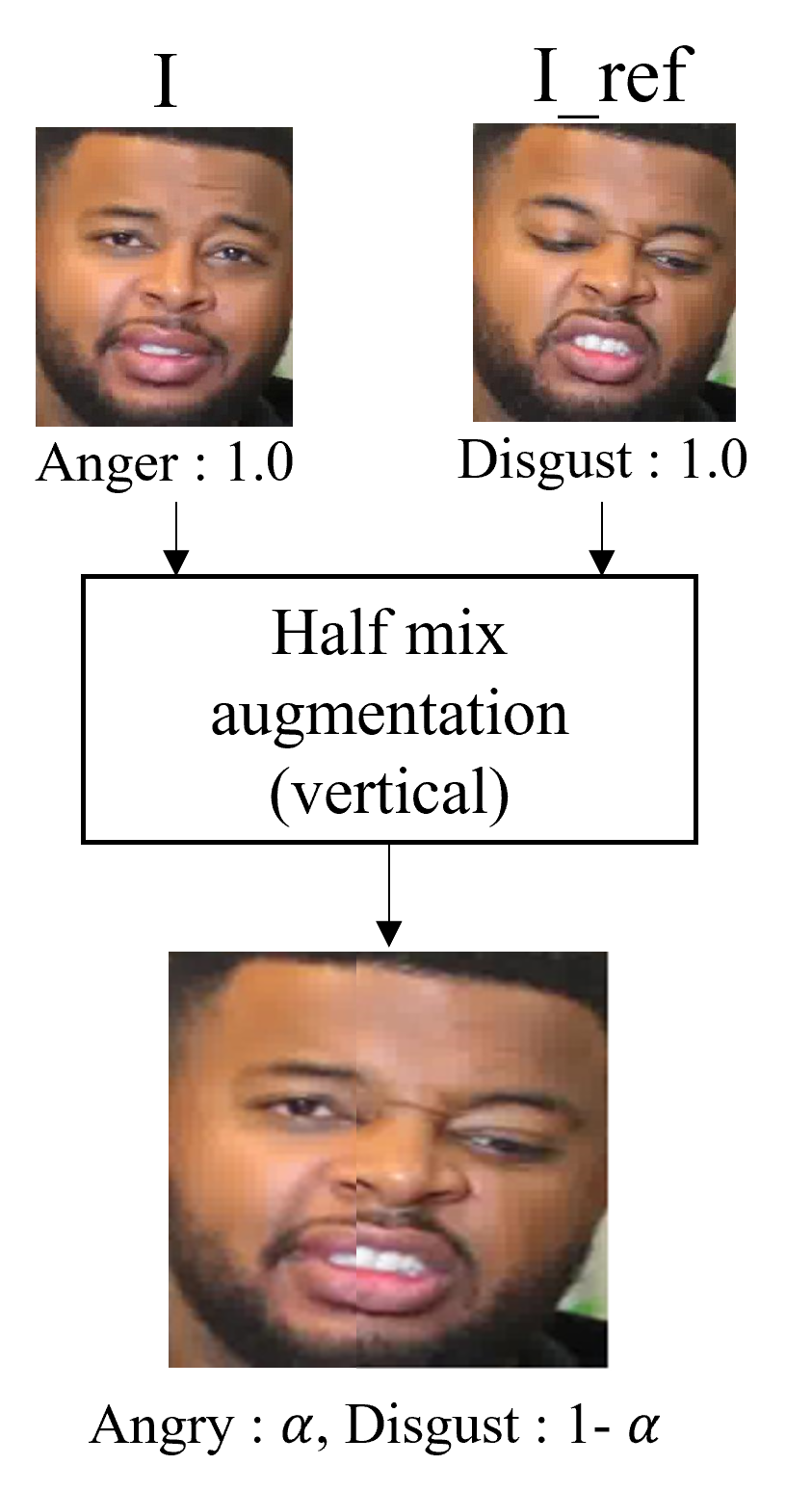}
         \caption{}
         \label{fig-(a)}
     \end{subfigure}
     \begin{subfigure}[b]{0.3\textwidth}
         \centering
         \includegraphics[width=\textwidth,height=7cm]{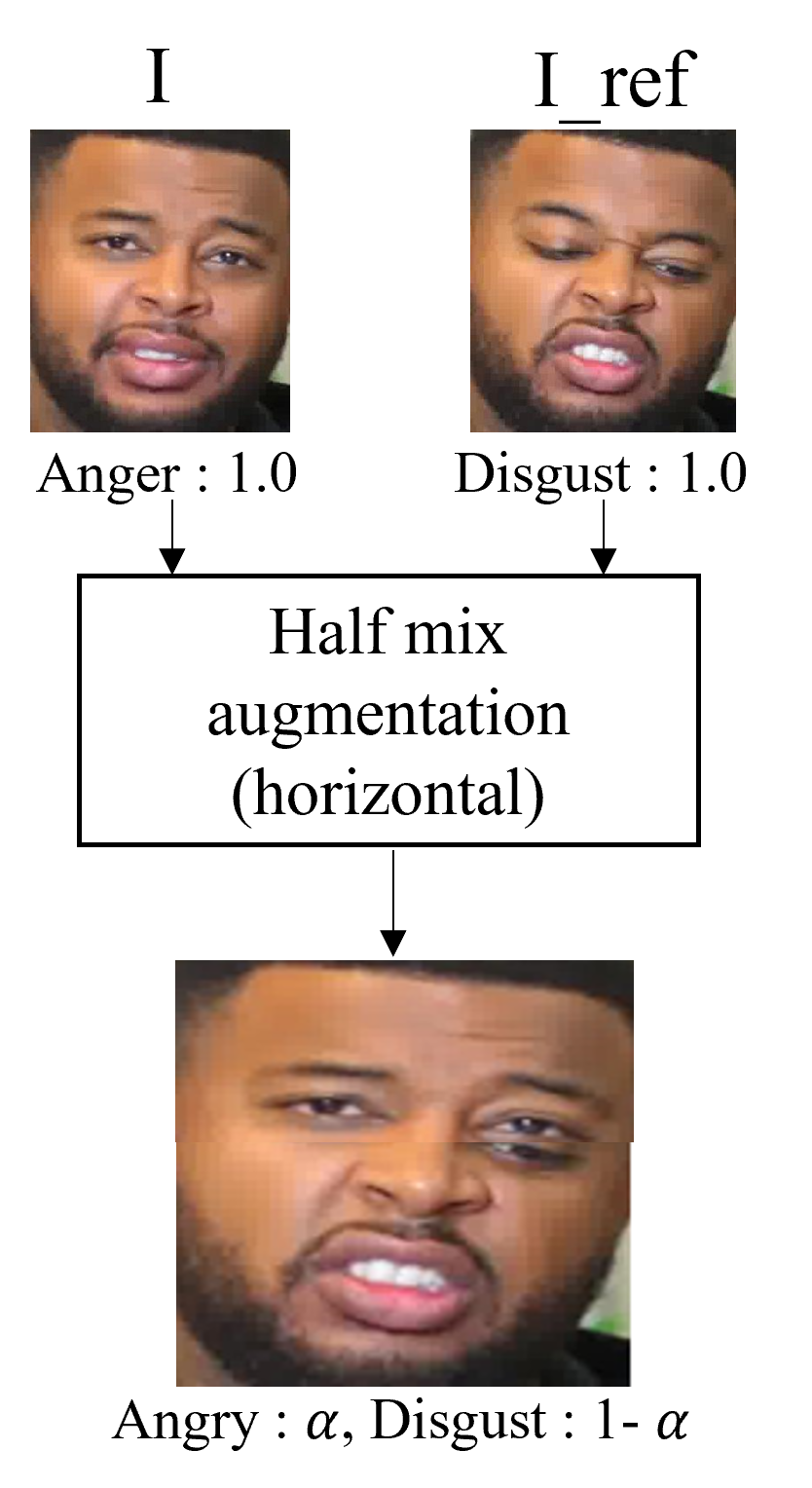}
         \caption{}
         \label{fig-(b)}
     \end{subfigure}

    \caption{(a) Half-mix jittering (vertical); (b) Half-mix jittering (horizontal);}
    \label{augmentation}
\end{figure}

We introduce half-mix jittering, a new data augmentation method suitable for face images. The half-mix jittering is based on the fact that humans can recognize facial emotions with only a half of the face. As shown in Figure \ref{augmentation}, in the training phase, we use the left partial face ($I_L$) or the right partial face ($I_R$) of the input $I$ and fill the rest with a reference image ($I_{ref}$) randomly selected within the batch. Similarly, the upper partial faces $I_T$ or the lower partial faces $I_D$ of the input $I$ are used. Let $x \in \mathbb{R}^{H \times W \times C}$ and $y$ denote a training image and label, respectively. To generate new training image $I(\tilde{x},\tilde{y})$ by combining input image $I(x,y)$ and reference image $I_{ref}(x,y)$. We define the combining operation as

\begin{equation}
    \tilde{x} = \mathbb{M} \odot I(x) + (1-\mathbb{M}) \odot I_{ref}(x)
\end{equation}
\begin{equation}
    \tilde{y} = \alpha \times I(y) + (1-\alpha) \odot I_{ref}(y) ,
\end{equation}
where $\mathbb{M} \in \{0,0.6\}^{H \times W \times C}$ denotes a binary mask indicating to crop from image, $\alpha \in \{0.4,0.6\}$. Since the Aff-Wild2 suffers from class imbalance as shown in the Table\ref{table_dataset_stat}, the minority classes such as Anger, Disgust, Fear, and Surprise are used as reference images when applying the half-mix jittering. When using the half-mix jittering, the following loss function is used
\begin{equation}
    \mathbb{L} = \alpha \times \mathbb{L}(pred, I(y)) + (1-\alpha) \times \mathbb{L}(pred, I_{ref}(y)),
\end{equation}
where $pred$ denotes a model output.

\section{Experiment results}

\subsection{Implementation details}

Our experiments were conducted on a desktop computer with the following specifications: implemented in PyTorch, Ubuntu 16.4 operating System, 128GB RAM, and GPU GeForce RTX 6000 with 22GB memory. For Aff-wild2, we used only cropped images provided by organizers and resize them to $224\times 224$ for network input. We adopted Swin transformer \cite{liu2021swin} Large model for image and audio stream of visual stream. For the visual stream, a combined model of S3D and Swin transformer Large was used. The details of the training parameters of each model are shown in Table \ref{hyp}. Based on the SGD optimizer as an optimizer, and the learning rate of $1e-3$ decreases by 10 from 40, 50, and 60 epochs in a total of 70 epochs using a multi-step scheduler. Also, the batch size is 32. The data was augmented by using the half-mix jittering, horizontal flip, AutoAug \cite{cubuk2018autoaugment} for the visual stream image. Also, horizontal flip, AutoAug, Random Crop for the video shot, and horizontal flip and AutoAug were used for the audio stream. We used only 10\% of the number of images for each class of Aff-Wild2 to try training multiple times due to insufficient computing resources. The image stream of the visual stream uses all frames for inference, and the video shot stream extracts and uses 16 frames every 1 second (30 frames). Audio stream is used to convert audio signal into mel-spectrogram every 2 seconds (60 frames).

\begin{table}

  \caption{Details of training parameters.}
  \centering
  \begin{tabular}{cccc}
    \toprule
    \multirow{2}{*}{Description} & \multicolumn{3}{c}{Model} \\ \cmidrule{2-4}
     & Visual & Temporal & Audio \\
    \midrule
    Epoch & \multicolumn{3}{c}{70} \\
    \midrule
    Batch size& \multicolumn{3}{c}{32}\\
    \midrule
    Optimizer& \multicolumn{3}{c}{SGD with momentum} \\
    \midrule
    Initial Learning rate& \multicolumn{3}{c}{1e-3} \\
    \midrule
    Scheduler& \multicolumn{3}{c}{Multi step} \\
    \midrule
    Loss & Soft Cross Entropy & Cross Entropy & Cross Entropy  \\
    \midrule
    \multirow{2}{*}{Augmentation}& Horizontal flip, halfmix jittering& Horizontal flip, random crop & Horizontal flip\\
    {} & Autoaug & Autoaug & Autoaug \\
    \midrule
    \# of input images & 1 & 16 & 1\\
    \midrule
    Sampling rate & Every frame & Every 1 sec & Every 2 sec\\

    \bottomrule
  \end{tabular}
  
  \label{hyp}
\end{table}

\subsection{Evaluation metric}

Evaluation metric used in CVPR ABAW 2022 Challenge is $F_1$ score. $F_1$ score is defined as weighted average of precision and recall. The $F_1$ score is defined as 
\begin{equation}
    F_{1}score = \frac{2 \times precision \times recall}{precision + recall}
\end{equation}

\subsection{Results}

Table \ref{table_results} shows the results from our experiment on the validation set of Aff-Wild2. Table \ref{table_results} contains the f1-score of each validation set of image, video shot, and audio 3 streams. It also includes score fusion results of all streams.

\begin{table}[h]

  \caption{Experimental results for Aff-Wild2 validation set.}
  \centering
  \begin{tabular}{cccccc}
    \toprule
    Model & Stream   & F1-score \\ \midrule
    Baseline & - & 23.00 \\
    \midrule
    \midrule
    Swin-Large & Visual  & 34.74  \\ 
    \midrule
    Swin-Large & Temporal  & 17.96\\ 
    \midrule
    Swin-Large & Audio  &18.99\\ 
    \midrule
    Swin-Large & Visual+Temporal    & 35.08\\
    \midrule
    Swin-Large & Visual+Temporal+Audio  &\textbf{35.71}\\ 

    \bottomrule
  \end{tabular}
  
  \label{table_results}
\end{table}

\section{Conclusion}

In this paper, we have exploited multi-modal data with a three-stream model, including cropped faces, multiple cropped faces, and audio obtained from the Aff-Wild2 data set, to solve the task of classifying eight facial expressions. Based on the recently introduced Swin-transformer, it showed better performance than the baseline, and performance improvement was achieved with the proposed half-mix jittering augmentation.

\bibliographystyle{unsrt}  
\bibliography{references}

\end{document}